# Robustness of Structured Data Extraction from Perspectively Distorted Documents

Hyakka Nakada
Data Technology Lab
Recruit Co.,Ltd.
Tokyo, Japan
hyakka_nakada@r.recruit.co.jp

Yoshiyasu Tanaka
Tokyo Office
Yamadentanaka Co., Ltd.
Tokyo, Japan
y-tanaka@yamadentanaka.com

*Abstract*—Optical Character Recognition (OCR) for data extraction from documents is essential to intelligent informatics, such as digitizing medical records and recognizing road signs. Multi-modal Large Language Models (LLMs) can solve this task and have shown remarkable performance. Recently, it has been noticed that the accuracy of data extraction by multi-modal LLMs can be affected when in-plane rotations are present in the documents. However, real-world document images are usually not only in-plane rotated but also perspectively distorted. This study investigates the impacts of such perturbations on the data extraction accuracy for the state-of-the-art model, Gemini-1.5-pro. Because perspective distortions have a high degree of freedom, designing experiments in the same manner as single-parametric rotations is difficult. We observed typical distortions of document images and showed that most of them approximately follow an isosceles-trapezoidal transformation, which allows us to evaluate distortions with a small number of parameters. We were able to reduce the number of independent parameters from eight to two, i.e. rotation angle and distortion ratio. Then, specific entities were extracted from synthetically generated sample documents with varying these parameters. As the performance of LLMs, we evaluated not only a character-recognition accuracy but also a structure-recognition accuracy. Whereas the former represents the classical indicators for optical character recognition, the latter is related to the correctness of reading order. In particular, the structure-recognition accuracy was found to be significantly degraded by document distortion. In addition, we found that this accuracy can be improved by a simple rotational correction. This insight will contribute to the practical use of multi-modal LLMs for OCR tasks.

## I. Introduction

Optical Character Recognition (OCR) is a technology that recognizes characters from document images [1]. It has many applications, including the digitization of medical records and recognition of road signs. In recent years, the introduction of deep learning has dramatically improved the accuracy of OCR [2]. In addition, powerful generative models are capable of understanding the context of a long document and providing answers to a variety of questions. Particularly, efforts have been reported to extract entities by leveraging a Large Language Model (LLM) [3,4]. Based on texts and their coordinates extracted by OCR, LLMs infer target entities. Multi-modal LLMs have been also utilized to extract entities by directly reading document images without OCR [5-8]. In other words, rather than recognizing characters just as written, it is now possible to interpret the context and summarize the desired entities with LLMs. Particularly, multi-modal-based methods have the advantage of reducing information loss because they can extract entities from images in a single pass without utilizing intermediate output of OCR. This study is concerned with entity extraction methods using multi-modal LLMs.

Scanners or smartphone cameras are usually used to capture document images. While a scanner produces a precise image that is faithful to the original document, smartphones generally produce distorted images. Document skew (i.e. in-plane rotation) is a well-known kind of distortion. In addition, perspective distortion is a more general kind of distortion. The plane of a camera film or CMOS sensor is usually not parallel to that of the document paper during hand-held shooting, which leads to perspective distortion. Even if they were parallel, in-plane rotation is inevitable. These perturbations are known to degrade the OCR performance, which has been an issue in the field of OCR analysis for several decades. In early studies, much attention was paid to the development of algorithms to detect and correct distortion in document images. The development of such algorithms paved the way for understanding the importance of correcting distortion for improving the OCR accuracy.

Several previous studies have investigated the effect of document rotation on OCR performance. Hull et al. proposed to realign rotation with connected component analysis so as to improve OCR performance [9]. Li et al. used wavelet decomposition and projection profile for detecting the rotation angle [10]. Recently, the application of deep learning has been proposed, leading to more sophisticated methods. Dobai et al. showed that the correction based on a neural network outperformed conventional methods [11]. Many other methods have been developed [12-15]. Several studies have been reported to resolve perspective distortion. Document boundary [16], text lines [17], character shape [18,19] can be used to correct distortion. In addition, Li et al. proposed the correction based on a neural network [20].

Without these corrections, accuracy degradation has recently been reported in entity extraction with multi-modal LLMs [21]. They generated synthetically rotated documents with varying angles of in-plane rotation and inferred the target entities to evaluate the dependence of the rotation angle on the entity-extraction accuracy.

As mentioned above, perspective distortion as well as rotation can occur in document images. More precisely, perspective distortion is the phenomenon that includes rotation. Because rotation occurs only when the image sensor and the plane of a document paper are coincidentally parallel, perspective distortion is more frequent and general. Therefore, in this study, we focus on evaluating the accuracy degradation against such distortions.

However, there are several challenges when handling perspective distortion. While rotation is described only by a single parameter (i.e. the rotation angle), perspective distortion is generally described by the homographic transformation [22]. This transformation has eight independent parameters as shown later. Thus, unlike the case of rotation in previous studies, grid-based methods are not feasible for evaluating the effect of perturbation due to combinatorial explosion. In addition, because the homographic transformation includes rotation, it becomes difficult to distinguish between rotational and other contributions. Thus, the dependency of the degree of perspective distortion on the entity-extraction accuracy is not easy to comprehensively evaluate.

In this paper, we develop a new experimental design for perspective distortion with a reduced number of parameters. We propose to utilize specific but practical distortions that are likely to occur when document images are captured. In practice, the manner of distortion in document images is biased due to constraints of the document and camera locations, and the photographer's intentions. Such restrictions can re-describe typical distortions in document images with fewer parameters. We evaluated the entity-extraction accuracy of the state-of-the-art multi-modal models, Gemini-1.5-pro [23] by synthesizing document images with these distortions.

## II. Previous Work: In-plane rotation

Recently, multi-modal LLMs have been applied to the task of entity extraction from document images [5-8]. They showed a great accuracy and have the potential to become standard for such a task instead of conventional OCR methods. A previous study applied multi-modal LLMs to entity extraction from rotated document images [21]. Specific entities were extracted from synthetically generated sample documents with varying angles of in-plane rotation to estimate the accuracy of entity extraction. As a result, they showed that document rotation significantly affects the performance of multi-modal LLMs. Accuracy degradation was observed when the rotation angle exceeded a specific threshold, several tens of degrees. Then, the accuracy recovered around the angle of 180°. The region below the threshold is called Safe In-plane Rotation Angles (SIPRA) [21]. Thus, they showed the importance of addressing document rotation, through rotation correction or by developing more robust model architectures and training approaches.

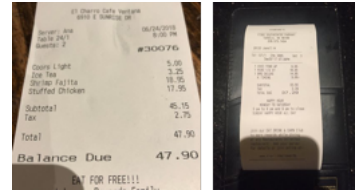

Fig. 1 Examples of shaded images. The shades of a smartphone can be seen partially.

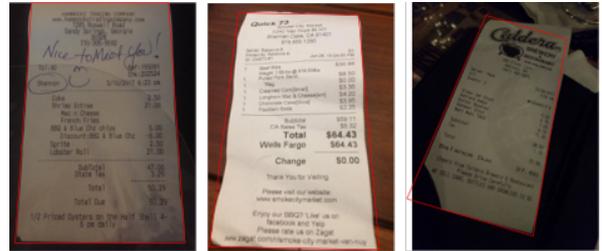

Fig. 2 Examples of receipt images. Red lines are showed as isosceles-trapezoidal guides for document areas.

Table 1 Ratios (%) of document area shapes. Exception means a complex shape other than quadrilateral, such as a shape strongly bended or folded.

| Rectangle | Isosceles trapezoid | Other quadrilateral | Exception |
|---|---|---|---|
| 35.5 | 60.5 | 1.5 | 2.5 |

## III. Methodology

### A. Generalization from Rotation to Perspective Distortion

When document images are captured with a smartphone, the subject is often set on a flat surface, such as a table. However, although it is desirable to capture images from directly above the paper, overhead lights tend to cast a shade on it, as shown in Fig. 1. For enhanced visibility, photographers often take pictures by moderately tilting their cameras. Therefore, we hypothesize that document areas are approximately isosceles-trapezoidally transferred, as shown in Fig. 2. We visually checked 200 images in the open-source receipt database [24] and classification for their shapes is listed in Table 1. While less than 36% images were found to be rectangles, more than 60% were almost isosceles trapezoids. Here, by redefining an isosceles trapezoid to include a rectangle, about 96% of the document images could be visually regarded as isosceles trapezoids. Although in-plane rotation can describe only the pattern of rectangles, isosceles-trapezoidal transformation is more versatile transformation for distortion in most document images. Therefore, in this study, we focus on isosceles-trapezoidal transformations, and evaluated the accuracy degradation against this type of distortion.

### B. Isosceles-trapezoidal Transformation

A schematic picture of isosceles-trapezoidal transformation is shown in Fig. 3 (a). The shape before transformation is called the original shape $OABC$. The parameters describing an isosceles trapezoid are the length of the upper bottom, that of the lower bottom, and the height of the transformed shape $OA'B'C'$.

We restrict ourselves to isosceles trapezoids that keep the area and height invariant to the original shape. Thus,

$$h = h_0, \quad (1)$$
$$(x + rx) \times h/2 = w_0 h_0 \quad (2)$$

are obtained. Here, $h_0$ is the height of the original shape, and $h$ is that of the transformed shape. $w_0$ is the width of the original shape. $x$ denotes the length of the upper bottom and $r > 0$ is the ratio of the lower bottom length to the upper bottom length,

$$r \equiv \frac{\text{(Lower bottom length)}}{\text{(Upper bottom length)}} = \frac{A'B'}{OC'} \quad (3)$$

$x$ and $h$ can be eliminated by $r$ according to Eqs. (1) and (2). Thus, the independent variables can be reduced to only $r$. Hereafter, $r$ will be referred to as the distortion ratio, the value of which takes around 1 when the shape is almost a rectangle. When $r \to \infty$, the length of the lower bottom approaches 0. On the other hand, the length of the upper bottom approaches 0 for $r \to 0$. That is, the shape will be triangular in these limits. The vertices of the transformed shape are explicitly calculated: $O(0,0)$, $A'((1-r)w_0/(1+r), h_0)$, $B'(w_0, h_0)$, and $C'(2w_0/(1+r), 0)$.

In addition, rotation is also combined with the above transformation, as shown in Fig. 3 (b). Specifically, the rightmost vertex of the upper bottom edge is rotated by an angle $\theta$. Here, $\theta$ is defined as positive clockwise. The vertices are explicitly transformed from $O(0,0), A(0, h_0), B(w_0, h_0)$, and $C(w_0, 0)$ into

$$O(0,0),$$
$$A''\left(\frac{(1-r)w_0}{1+r}\cos\theta - h\sin\theta, \frac{(1-r)w_0}{1+r}\sin\theta + h\cos\theta\right),$$
$$B''(w_0\cos\theta - h_0\sin\theta, w_0\sin\theta + h_0\cos\theta),$$
$$C''\left(\frac{2w_0}{1+r}\cos\theta, \frac{2w_0}{1+r}\sin\theta\right). \quad (4)$$

The homographic transformation can map between two planar projections of an image [22]. This is described by

$$s\begin{pmatrix}x'\\y'\\1\end{pmatrix} = H\begin{pmatrix}x\\y\\1\end{pmatrix} \equiv \begin{pmatrix}h_{11} & h_{12} & h_{13}\\h_{21} & h_{22} & h_{23}\\h_{31} & h_{32} & h_{33}\end{pmatrix}\begin{pmatrix}x\\y\\1\end{pmatrix}. \quad (5)$$

While $(x, y)$ is the coordinates in the original shape, $(x', y')$ denotes that in the transformed shape. $H$ is a projection matrix, and $h_{11}, \dots, h_{33}$, and $s$ are the parameters of the homographic transformation. Equation (5) is rewritten in the following form

$$x' = \frac{h_{11}x + h_{12}y + h_{13}}{h_{31}x + h_{32}y + h_{33}}$$
$$y' = \frac{h_{21}x + h_{22}y + h_{23}}{h_{31}x + h_{32}y + h_{33}} \quad (6)$$

by eliminating $s$. Although there are apparently nine parameters, $h$ has arbitrariness regarding magnitude due to division. Thus, the effective degrees of freedom are eight. Shifting from $OABC$ to $OA''B''C''$ (Eq. (4)) uniquely determines the equivalent projection matrix

$$H = \begin{pmatrix}\cos\theta & \dfrac{(1-r)w_0\cos\theta - (1+r)h_0\sin\theta}{2rh_0} & 0\\ \sin\theta & \dfrac{(1-r)w_0\sin\theta + (1+r)h_0\cos\theta}{2rh_0} & 0\\ 0 & -\dfrac{(1-r)(1+r)}{2rh_0} & \dfrac{1+r}{2}\end{pmatrix}. \quad (7)$$

Here, there are only two parameters for the total transformation: the rotation angle $\theta$ and the distortion ratio $r$. In the limit of $r \to 1$, that is without distortion, Eq. (7) reproduces to an in-plane rotation matrix. On the other hand, when $\theta \to 0$, the isosceles-trapezoidal transformation in Fig. 3 (a) is obtained. Our formalism can describe the multifaceted transformation between these limits.

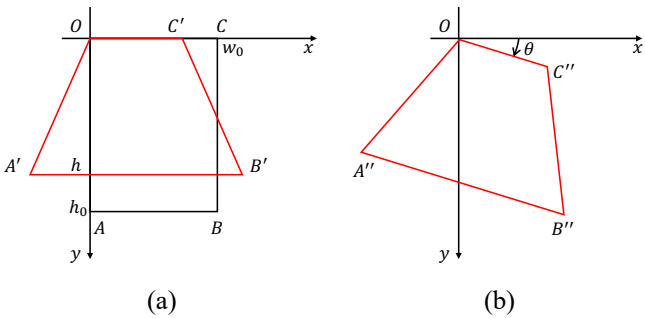

(a)  (b)

Fig. 3 (a) Schematic picture of isosceles-trapezoidal transformation. A Cartesian coordinate system is set along the sides $OA$ and $OC$ of a rectangle $OABC$, which represents an original document shape. The transferred shape is depicted as an isosceles trapezoid $OA'B'C'$. (b) Then, rotation by angle $\theta$ is performed with $O$ the rotation center.

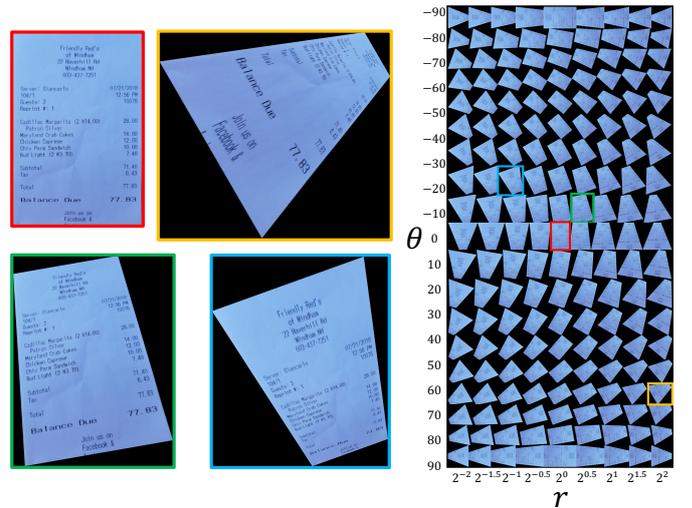

Fig. 4 Original image (red frame) is transferred by Eq. (7). For every values of $\theta$ and $r$, transferred documents are enumerated matrix-like. Their several examples are also shown (yellow, green, and blue frames).

## IV. Experiment

### A. Data sets

Ten images with almost no distortion or rotation in the open-source receipt database [24] were selected as original images and subjected to our transformation. The pair of parameters $\theta, r$ is set in the grid of $\theta = [-90, -80, ..., 0, ..., 90]$, $r = [2^{-2}, 2^{-1.5}, ..., 2^0, ..., 2^2]$, as shown in Fig. 4. In other words, for each original shape, $19 \times 9 = 171$ transformed images are generated according to Eq. (7). Thus, the total number of verification data is 1710.

### B. Multi-Modal LLM

We used the prompt shown in Fig. 5 to extract eight entities: *vendor*, *date*, *list item*, *subtotal*, *tax*, *total*, *payment*, and *change*. It is a schema-type prompt for JSON. For simplicity, we gave the following conditions. Items with no price or zero price are excluded, and the price of the item should not be for a unit but for the entire quantity. *Vendor* is case sensitive. In addition, we defined that *total* is the sum of *subtotal* and *tax*. As a multi-modal LLM, we use Gemini-1.5-Pro-002 with temperature 1.0, max output tokens 8192, and Top-P 0.95.

### C. Evaluation

An example of LLM outputs responded by the prompt in Fig. 5 is shown in Fig. 6. By referring to keys, we extract the corresponding value of *vendor*, *date*, *list item*, *subtotal*, *tax*, *total*, *payment*, and *change*. Each score is estimated in the following. *Subtotal*, *tax*, *total*, *payment*, and *change* have numerical values. As shown in Fig. 6, while the true value of *subtotal* is 71.40, the predicted value is 71.4. Such a mismatch was frequently observed. This is probably because the LLM eliminates unnecessary information for summarization. Thus, the scores for these entities are set to 1 if a predicted value is numerically equal to a true value (e.g. $71.40 = 71.4$), and 0 otherwise.

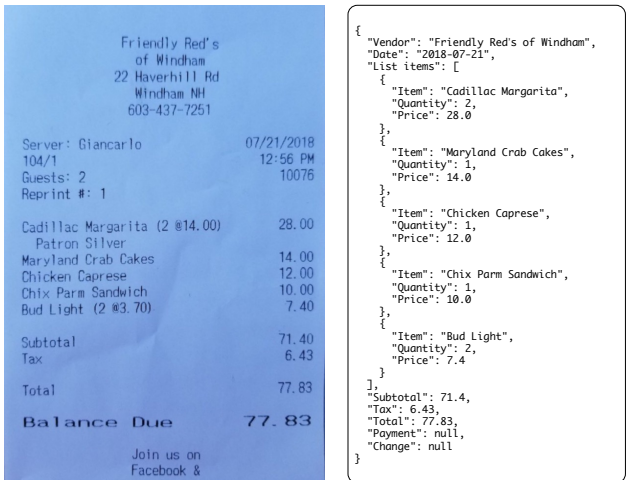

Fig. 5 Prompt into Gemini-1.5-pro for entity extraction.

Fig. 6 Input document image (left) and LLM output (right). JSON data is outputted. By referring to keys, corresponding values are obtained. For the entity which an LLM cannot extract, the corresponding value is padded by null.

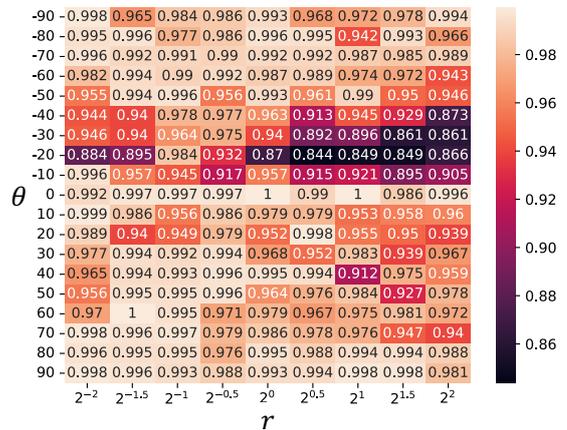

Fig. 7 Average scores are estimated by averaging all of scores for *vendor*, *date*, *list item*, *subtotal*, *tax*, *total*, *payment*, and *change*. Darker areas mean the worse scores.

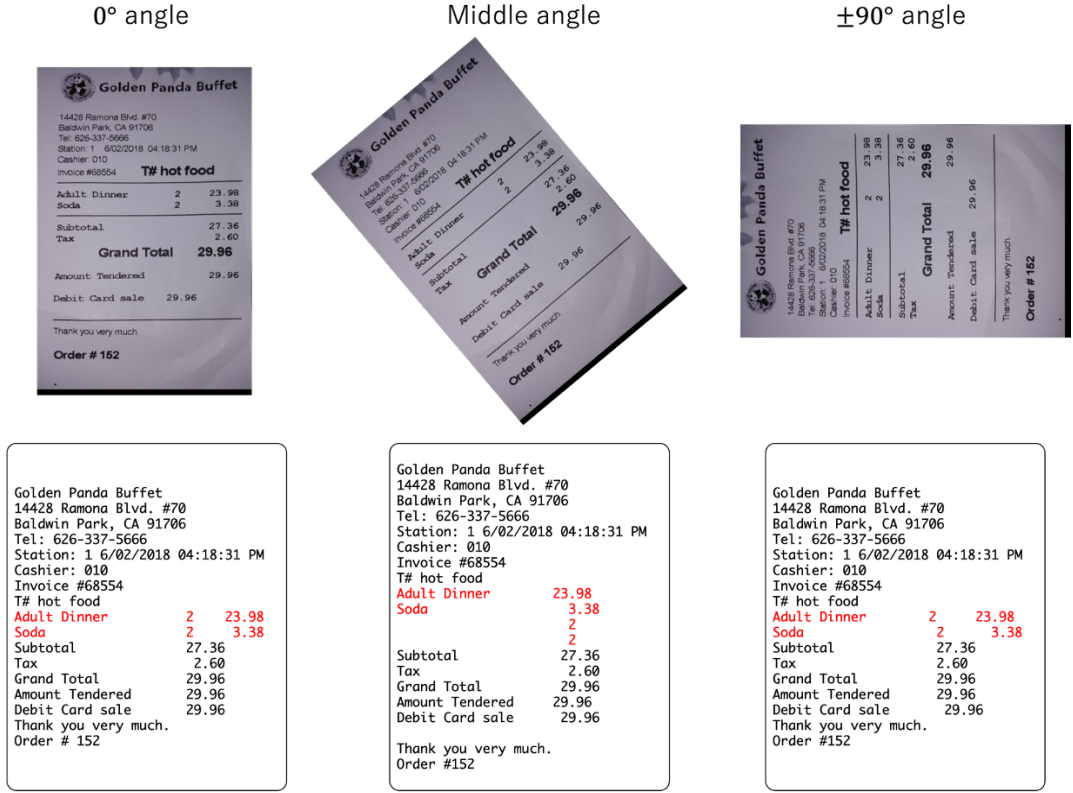

Fig. 8 Document images rotated by typical angles. In lower side, full-texts extracted by a prompt of Fig. 12. In a middle angle, the quantity values are shifted from the original lines.

For *vendor* and *date*, their values are relatively lengthy strings. As scores, we adopt a Jaro-Winkler distance between a true value and predicted value. Finally, because *list item* is sorted information, we estimate the score based on a reading order. In other words, from top to bottom, a true item and predicted item are compared. Individually, scores for quantity and price are calculated in the same manner of entities with numerical values, such as *subtotal*. Scores for item name are calculated by the Jaro-Winkler distance, like *vendor*. *List item* scores are estimated by averaging these scores.

## V. Results

### A. Entity Extraction Accuracy

The average scores for all indicators are summarized in Fig. 7. First, the results are explained with respect to rotation angle $\theta$. When the angle is $\theta = 0°, \pm 90°$, relatively high scores are achieved compared to other angles. The middle angles gave the accuracy degradation, which was consistent with the results of the previous study [21]. In other words, the valleys of average scores are resolved at $\theta = \pm 90°$. Nevertheless, while the severe decrease appears in the negative angles (i.e. counterclockwise rotated image), the change of the accuracy is relatively moderate in the positive.

With respect to distortion ratio $r$, the behavior of the score is more complex. At $\theta = 0°, \pm 90°$, the entity-extraction accuracy was robust against the values of $r$. On the other hand, the middle region between such angles showed the degradation with large or small $r$. In this way, it was confirmed that performing not only rotation but also perspective distortion generates a rich phenomenon. Because many real-world document images include the latter perturbation, this result has great applicability. If $\theta = 0°, \pm 90°$, i.e. rows or columns are aligned as shown in Fig. 8, the degree of distortion appears to have little effect on the accuracy. Therefore, for practical use, it is expected that highly accurate entity-extraction can be achieved simply by performing rotation correction without the need for perspective-distortion correction, which is generally more laborious than the rotation correction.

### B. Considerations for Detailed Scores

For further discussion, we observe the score for each entity. The scores for *vendor* and *date* were estimated by the Jaro-Winkler distance between a true string and predicted string. Thus, they are considered to reflect the power of character recognition. As shown in Fig. 9, a slight asymmetry of the scores can be confirmed with respect to $r$.

Such an asymmetry may exist because the information of *vendor* and *date* is usually written on the upper side of receipts. If $r \gg 1$, i.e. the upper bottom is shrunk, the difficulty of character recognition generally increases. For example, the yellow-framed image in Fig. 4 has small font sizes in *vendor* and *date*. Nevertheless, Fig. 9 (a) shows that almost all scores are more than 0.98, the accuracy of character recognition was found to be largely unaffected except under excessive distortion.

On the other hand, the scores for *subtotal*, *tax*, *total*, *payment*, and *change* reflect the power of structure recognition, which is related to reading order. As mentioned above, the LLM has a great ability of character recognition. Especially, we found that number characters were almost perfectly recognized, as the *date* score in Fig. 9 (b). Thus, mainly, misunderstanding the order relationship between a key and value is considered to contribute to degradation of the scores of *subtotal*, *tax*, *total*, *payment*, and *change*.

In addition, the *list item* score reflects the power of reading order by definition as mentioned in Chapter IV C. While *subtotal*, *tax*, *total*, *payment*, and *change* have each one value, *list item* is comprised of multiple keys and values. For example, as shown in Fig. 6, five pairs of a menu and price are required to extract from top to bottom. Here, "Patron Silver" without prices should be discarded. Their scores were found to decrease significantly compared with those of *vendor* and *date*, as shown in Fig. 10. Note that the degradation of the *change* score is relatively moderate. This is because *change* was not often written in the receipts and many of their true values were null.

Particularly, the rotation angle has an impact on these degradations. In Fig. 10 (a)-(e), the scores are slightly degraded in the positive angles and the severe decrease appears in the negative. The score of (f) *list item* shows the severe decrease in both regions. Two valleys exist around $\theta = -20°, 70°$ and the example of extraction results is shown in Fig. 11. This tendency becomes grater for large values of $r$. Whereas some prices were misassigned and items were partially lost at such angles, the other angles show successful extraction.

The reason for this degradation can be considered as follows. For example, in a rotated document as shown in the upper part of Fig. 8, the latitude of a key location was shifted far from that of the corresponding value's location. As a result, different values were sometimes assigned, leading to the low scores. By using the prompt in Fig. 12 for extracting all the sentences, it was found that the LLM did not accurately capture their row structure, as shown in the red texts of Fig. 8. At $\theta = 0°$, an item, quantity, and price were written in the same rows. On the other hand, at a middle angle, they were often observed to be scattered over several rows. Such destruction may lead to misunderstanding the order relationship between a key and value. In fact, according to the results of extraction shown in Fig. 11, the prices of some items have been misassigned.

The destruction resolves at $\theta = \pm 90°$ and the high accuracy is achieved again. This recovery may be reasonable because LLMs are supposed to learn plentiful document data, which are usually documents vertically or horizontally scanned. In addition, the order relationship may have been successfully understood because the information is aligned not in a row, but in a column.

Finally, we remark on the dependency of the distortion ratio $r$ on the entity-extraction accuracy. Largely, this distortion has a negative effect on the accuracy in the middle angles, implying that structure recognition is degraded more than character recognition. In addition, while the effect of magnifying the upper bottom is moderate, magnifying the lower bottom causes significant deterioration except for the *change* score. Considerations for this asymmetry will be the subject of future work. Again, the above degradation occurs mainly among the middle angles of rotation, little deterioration was observed near $\theta = 0°, \pm 90°$.

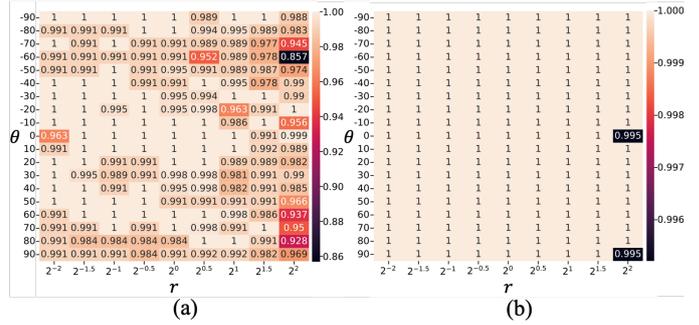

Fig. 9 (a) *Vendor* scores and (b) *date* scores are enumerated. Darker areas mean the worse scores.

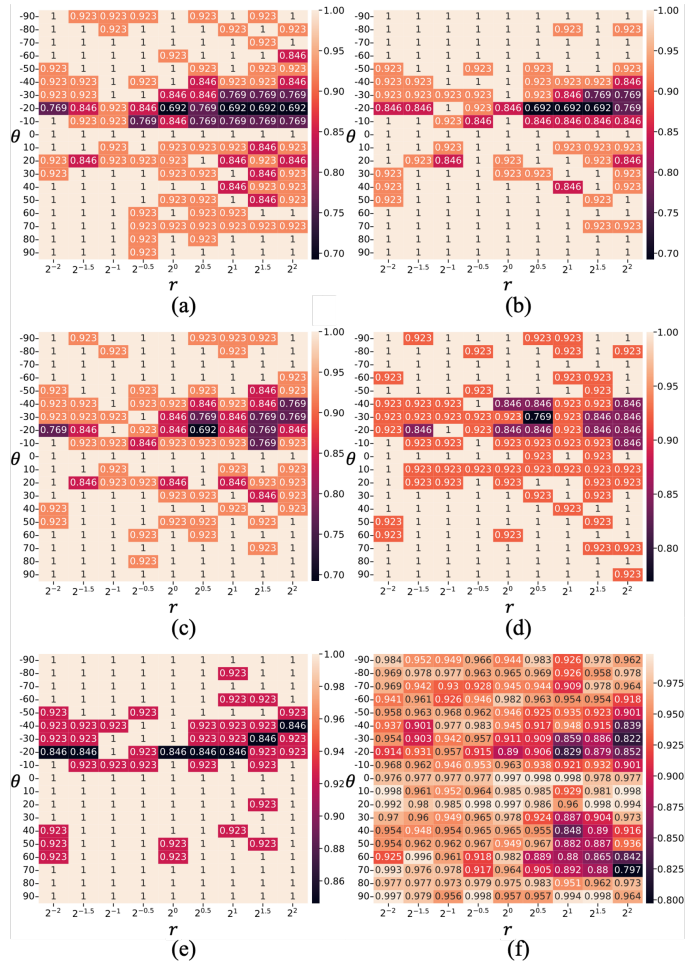

Fig. 10 (a) *Subtotal*, (b) *tax*, (c) *total*, (d) *payment*, (e) *change*, and (f) *list item* scores are enumerated. Darker areas mean the worse scores.

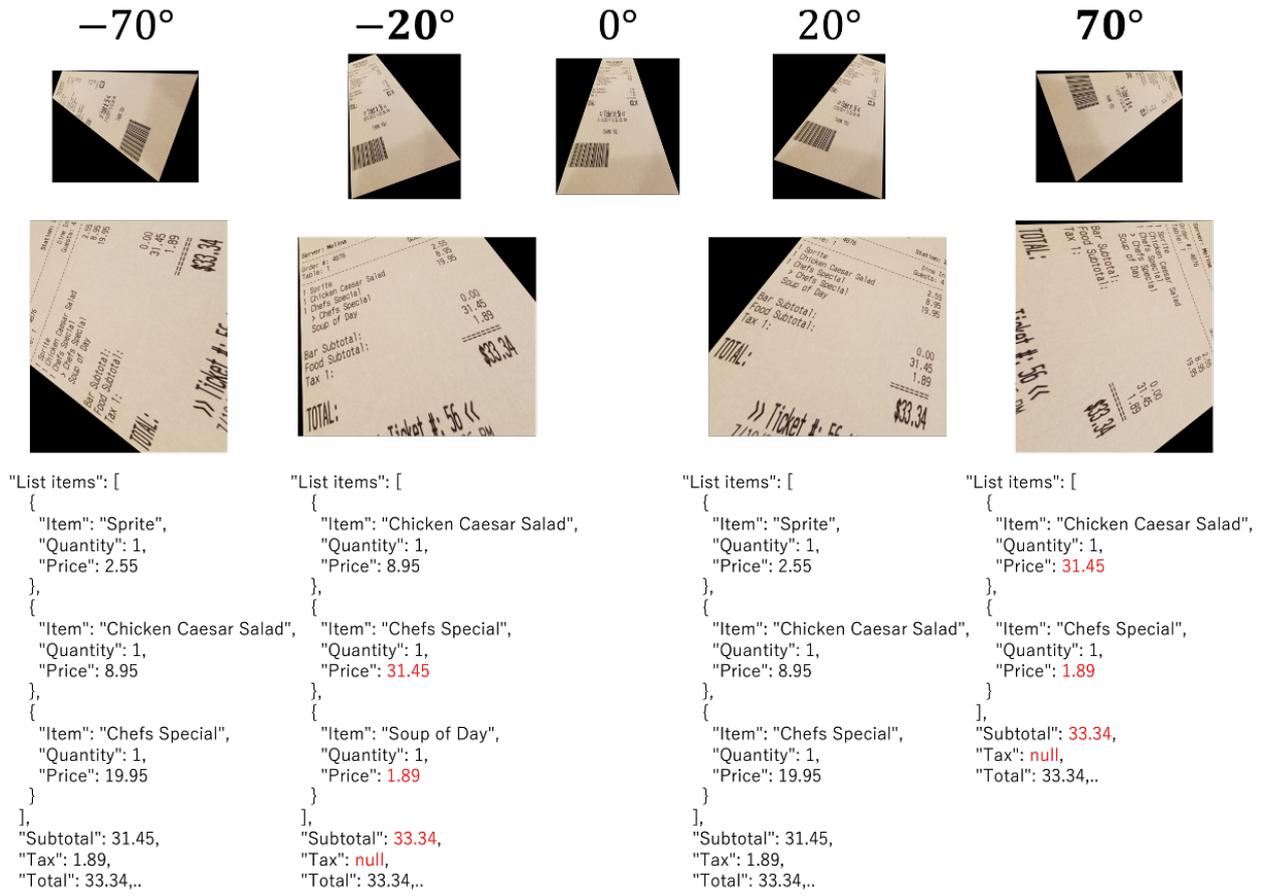

Fig. 11 Under $r = 4$, several rotated images are shown. In addition, their cropped images and the partial results of entity extraction are enumerated.

Fig. 12 Prompt into Gemini-1.5-pro for full-text extraction.

## VI. CONCLUSION

When multi-modal LLMs are applied to data extraction from documents, the entity extraction accuracy has been reported to be affected by in-plane rotation of documents. However, real-world document images are usually not only in-plane rotated but also perspectively distorted. This study investigates their impacts on the entity extraction accuracy for the state-of-the-art multi-modal LLMs, Gemini-1.5-pro. Designing experiments in the same manner as single-parametric rotations is challenging because perspective distortions have a high degree of freedom. Thus, we eliminated the degree of freedom for insignificant parameters by observing typical distortions of document images, to find that most of them approximately follow an isosceles-trapezoidal transformation. This type of distortion has only two parameters, i.e. rotation angle $\theta$ and distortion ratio $r$. Then, specific entities were extracted from synthetically generated sample documents with varying these parameters.

As the performance of LLMs, we evaluated not only a character-recognition accuracy but also a structure-recognition accuracy. Whereas the former represents the classical indicators for optical character recognition, the latter relates with the correctness of reading order. Especially, the structure-recognition accuracy was found to be significantly degraded due to document transformation. In addition, we found that only correcting rotation is sufficient to improve this accuracy. This insight will contribute to the practical use of multi-modal LLMs for OCR tasks.

Future works are discussed. Besides perspective distortions, document images generally contain noises, such as character blur. LLMs can be used to compensate for these character defects based on the context of the document. Thus, we would like to evaluate the robustness of LLM performance against such a noise. In addition, we proposed a new experimental design for perspective distortion in this study. Our method is expected to augment high-quality training data for OCR because our transformation can automatically generate data that mimics real document images, as shown in Table 1. It can be applied to general tasks of object detection [25] rather than OCR.

ACKNOWLEDGMENT

The authors wish to thank Masakazu Yakushiji, Takashi Egami, Rinka Fukuji, and Marika Kubota for their advice. We also thank Ryo Takahashi and Viviane Takahashi and our anonymous reviewers for their comments.

REFERENCES

[1] S. Mori, H. Nishida, and H. Yamada, "Optical character recognition," John Wiley & Sons, Inc., 1999.

[2] N. Subramani, A. Matton, M. Greaves, and A. Lam, "A survey of deep learning approaches for ocr and document understanding," arXiv preprint arXiv:2011.13534, 2020.

[3] V. Perot, K. Kang, F. Luisier, G. Su, X. Sun, et al., "LMDX: Language model-based document information extraction and localization," arXiv preprint arXiv:2309.10952, 2023.

[4] F. Loukil, S. Cadereau, H. Verjus, M. Galfre, K. Salamatian, et al., "LLM-centric pipeline for information extraction from invoices," In Proceedings of the 2nd International Conference on Foundation and Large Language Models (FLLM), 2024, pp. 569-575.

[5] J. Ye, A. Hu, H. Xu, Q. Ye, M. Yan, et al., "mplug-docowl: Modularized multimodal large language model for document understanding," arXiv preprint arXiv:2307.02499, 2023.

[6] D. Wang, N. Raman, M. Sibue, Z. Ma, P. Babkin, et al., "Docllm: A layout-aware generative language model for multimodal document understanding," arXiv preprint arXiv:2401.00908, 2024.

[7] Y. Liu, B. Yang, Q. Liu, Z. Li, Z. Ma, et al., "Textmonkey: An ocr-free large multimodal model for understanding document," arXiv preprint arXiv:2403.04473, 2024.

[8] Y. Liu, Z. Li, M. Huang, B. Yang, W. Yu, et al., "Ocrbench: on the hidden mystery of ocr in large multimodal models," Science China Information Sciences, vol. 67, no. 12, pp. 220102, December 2024.

[9] J. J. Hull, "Document image skew detection: Survey and annotated bibliography," Document Analysis Systems II, 1998, pp. 40-64.

[10] S. Li, Q. Shen, and J. Sun, "Skew detection using wavelet decomposition and projection profile analysis," Pattern Recognition Letters, vol. 28, no. 5, pp. 555-562, April 2007.

[11] L. Dobai and M. Teletin, "A document detection technique using convolutional neural networks for optical character recognition systems," In Proceedings of the 27th European Symposium on Artificial Neural Networks, 2019, pp. 547-552.

[12] V. N. M. Aradhya, G. H. Kumar, and P. Shivakumara, "An accurate and efficient skew estimation technique for south indian documents: a new boundary growing and nearest neighbor clustering based approach," Int. J. Robot. Autom., vol. 22, no. 4, pp. 272-280, September 2007.

[13] A. A. Mascaro, G. D. Cavalcanti, and C. A. Mello, "Fast and robust skew estimation of scanned documents through background area information," Pattern Recognition Letters, vol. 31, no. 11, pp. 1403-1411, August 2010.

[14] M. Shafii, "Optical character recognition of printed persian/arabic documents," Ph.D. thesis, 2014.

[15] A. Al-Khatatneh, S. A. Pitchay, and M. Al-qudah, "A review of skew detection techniques for document," In Proceedings of the 17th UKSim-AMSS International Conference on Modelling and Simulation (UKSim), 2015, pp. 316-321.

[16] P. Clark and M. Mirmehdi, "Location and recovery of text on oriented surfaces," In Proceedings of the SPIE Document Recognition and Retrieval VII, 1999, pp. 267-277.

[17] P. Clark and M. Mirmehdi, "Estimating the orientation and recovery of text planes in a single image," In Proceedings of the 12th British Machine Vision Conference (BMVC2001), 2001, pp. 421-430.

[18] J. Liang, D. DeMenthon, and D. Doermann, "Geometric rectification of camera-captured document images," IEEE Transactions on Pattern Analysis and Machine Intelligence, vol. 30, no. 4, pp. 591-605, April 2008.

[19] S. Lu, B. M. Chen, and C. Ko, "Perspective rectification of document images using fuzzy set and morphological operations," Image and Vision Computing, vol. 23, no. 5, pp. 541-553, May 2005.

[20] X. Li, B. Zhang, P. V. Sander, and J. Liao, "Blind geometric distortion correction on images through deep learning," In Proceedings of the IEEE/CVF Conference on Computer Vision and Pattern Recognition (CVPR), 2019, pp. 4850-4859.

[21] A. Biswas and W. Talukdar, "Robustness of structured data extraction from in-plane rotated documents using multi-modal large language models (LLM)," Journal of Artificial Intelligence Research, vol. 4, no. 1, pp. 176-195, March 2024.

[22] Y. Luo, X. Wang, Y. Liao, Q. Fu, C. Shu, et al., "A review of homography estimation: Advances and challenges," Electronics, vol. 12, no. 24, pp. 4977, December 2023.

[23] G. Team, P. Georgiev, V. I. Lei, R. Burnell, L. Bai, et al., "Gemini 1.5: Unlocking multimodal understanding across millions of tokens of context," arXiv preprint arXiv:2403.05530, 2024.

[24] ExpressExpense, "Free receipt images," https://expressexpense.com/blog/free-receipt-images-ocr-machine-learning-dataset/, 2020.

[25] K. Wang, B. Fang, J. Qian, S. Yang, X. Zhou, and J. Zhou, "Perspective transformation data augmentation for object detection," IEEE Access, vol. 8, pp. 4935-4943, December 2020.